\documentclass{article}

\usepackage{microtype}
\usepackage{graphicx}
\usepackage{subfigure}
\usepackage{booktabs} 

\usepackage{hyperref}

\usepackage[accepted]{icml2024}

\usepackage{amsmath}
\usepackage{amssymb}
\usepackage{mathtools}
\usepackage{amsthm}

\usepackage[capitalize,noabbrev]{cleveref}

\theoremstyle{plain}

\theoremstyle{definition}

\theoremstyle{remark}

\usepackage[textsize=tiny]{todonotes}

\usepackage{color}
\usepackage{arydshln}
\usepackage{bm}
\usepackage{array}
\usepackage{comment}
\usepackage{caption}
\usepackage{colortbl} 
\usepackage{multirow}
\usepackage{makecell}
\usepackage{wrapfig}

\usepackage[normalem]{ulem}
\usepackage{arydshln} 
\usepackage{enumitem}

\usepackage{pifont}

\usepackage{tabularray}
\usepackage{paralist}
\usepackage{makecell}
\usepackage{colortbl}
\usepackage{etoolbox}

\usepackage{rotating}
\usepackage{enumerate}
\usepackage{array}
\usepackage{tabularx}
\usepackage{arydshln}
\usepackage{multirow}
\usepackage{xspace}
\usepackage{utfsym}
\usepackage{tcolorbox}
\usepackage{adjustbox}

\renewcommand{\footnotesize}{\fontsize{8pt}{11pt}\selectfont}

\definecolor{myblue}{RGB}{52,218,247}
\definecolor{myred}{RGB}{255,90,90}
\definecolor{mypink}{RGB}{239,43,159}
\definecolor{myupdate}{RGB}{254,243,222}
\definecolor{myfrozen}{RGB}{237,255,255}
\definecolor{ired}{RGB}{229,72,72}
\definecolor{igreen}{RGB}{44,222,139}
\definecolor{ired}{RGB}{247,142,142}
\definecolor{greeni}{RGB}{166,247,166}
\definecolor{a3c}{rgb}{0.00,0.50,0.50}
\definecolor{nmgray}{RGB}{229,229,229}
\definecolor{bluei}{RGB}{218,232,252}

\newcommand{\videologo}{\raisebox{-4pt}{\includegraphics[width=1.8em]{illus/video-icon1.pdf}}\xspace\xspace}
\newcommand{\stsglogo}{\raisebox{-3pt}{\includegraphics[width=2.8em]{illus/stsg-icon1.pdf}}\xspace\xspace}

\newcommand{\circlenum}[1]{%
    \resizebox{!}{0.8em}{%
        \tikz[baseline=(char.base)]{
            \node[shape=circle, fill=black, inner sep=0.8pt, text=white] (char) {#1};
        }%
    }%
}

\newtcolorbox{mybox}[2][]{
width=\columnwidth,
colback = nmgray!75!white, 
colframe = nmgray!75!white, 
boxsep=0pt,left=9pt,right=10pt,top=0pt,bottom=0pt,
fontupper=\linespread{0.9}\selectfont,
title=#2,#1}

\icmltitlerunning{Video-of-Thought: Step-by-Step Video Reasoning from Perception to Cognition}

\begin{document}

\twocolumn[

\icmltitle{\textit{Video-of-Thought}: Step-by-Step Video Reasoning from Perception to Cognition}



\icmlsetsymbol{equal}{*}

\begin{icmlauthorlist}
\icmlauthor{Hao Fei}{nus}
\icmlauthor{Shengqiong Wu}{nus}
\icmlauthor{Wei Ji}{nus}
\icmlauthor{Hanwang Zhang}{ntu}
\icmlauthor{Meishan Zhang}{sch}
\icmlauthor{Mong Li Lee}{nus}
\icmlauthor{Wynne Hsu}{nus}
\end{icmlauthorlist}

\icmlaffiliation{nus}{National University of Singapore, Singapore}
\icmlaffiliation{ntu}{Nanyang Technological University, Singapore}
\icmlaffiliation{sch}{Harbin Institute of Technology (Shenzhen), China}

\icmlcorrespondingauthor{Meishan Zhang}{zhangmeishan@hit.edu.cn}


\vskip 0.3in
]



\printAffiliationsAndNotice{}  

\begin{abstract}
Existing research of video understanding still struggles to achieve in-depth comprehension and reasoning in complex videos, primarily due to the under-exploration of two key bottlenecks: \emph{fine-grained spatial-temporal perceptive understanding} and \emph{cognitive-level video scene comprehension}.
This paper bridges the gap by presenting a novel solution.
We first introduce a novel video Multimodal Large Language Model (MLLM), \textbf{MotionEpic}, which achieves fine-grained pixel-level spatial-temporal video grounding by integrating video spatial-temporal scene graph (STSG) representation.
Building upon MotionEpic, we then develop a Video-of-Thought (\textbf{VoT}) reasoning framework. 
VoT inherits the Chain-of-Thought (CoT) core, breaking down a complex task into simpler and manageable sub-problems, and addressing them step-by-step from a low-level pixel perception to high-level cognitive interpretation.
Extensive experiments across various complex video QA benchmarks demonstrate that our overall framework strikingly boosts existing state-of-the-art. 
To our knowledge, this is the first attempt at successfully implementing the CoT technique for achieving human-level video reasoning, where we show great potential in extending it to a wider range of video understanding scenarios.
Project is open at \url{https://haofei.vip/VoT}.
\end{abstract}

\section{Introduction}
\label{Introduction}

Enabling learning models to accurately interpret video data is one of the most paramount goals in the relevant community. 
In the current research, while there has been extensive exploration into building models for video action and dynamics recognition \cite{lei2018tvqa,bertasius2021space}, mostly they fall prey to the type of straightforward perceptual-level understanding, i.e., for simple videos \cite{zolfaghari2018eco,lin2019tsm}.
And there remains a significant gap in research concerning comprehending and reasoning about complex videos in depth, an imperative capability urgently needed in real-world applications. 
Compared to shallow video perception, reasoning about complex videos poses greater challenges: it demands not only an intricate understanding of the video's spatiotemporal characteristics \cite{caballero2017real}, but also a profound grasp of the underlying implications behind pixels.

\begin{figure}[!t]
\centering
\includegraphics[width=1\columnwidth]{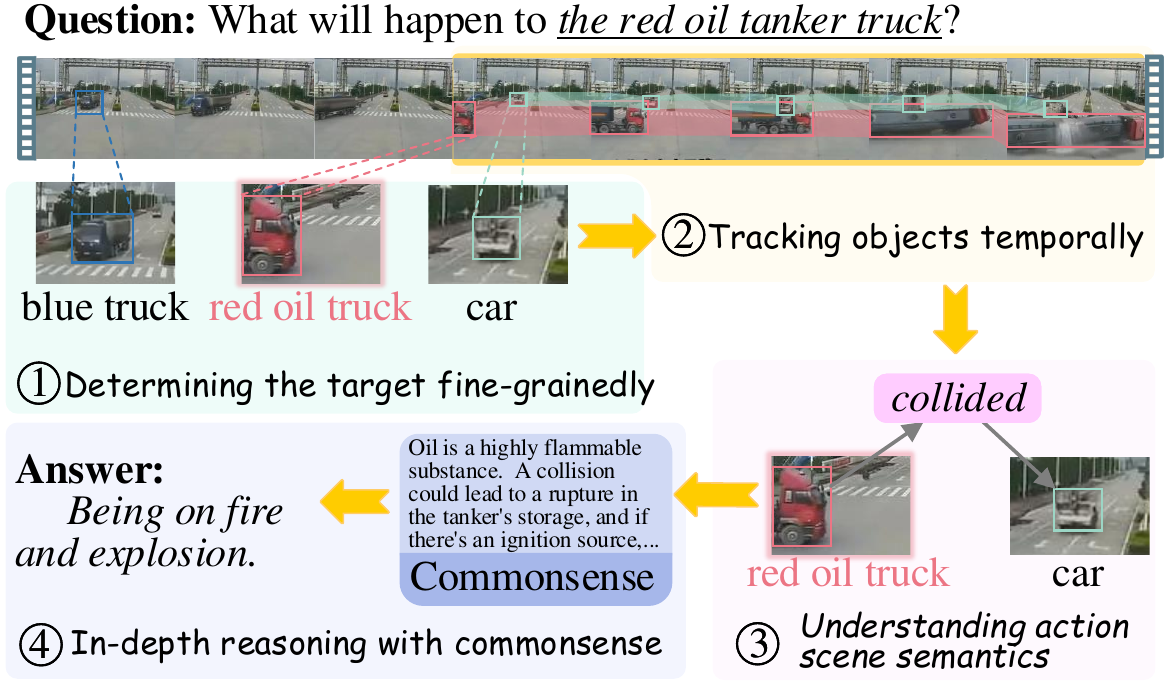}
\caption{
Human-like video reasoning intuitively follows a multi-step procedure, from lower-level perceptive fine-grained pixel grounding and tracking, to higher-level cognitive action scene semantics understanding.
}
  \vspace{-5mm}
\label{fig:intro}
\end{figure}

Drawing from human cognition patterns, we mark that reasoning about videos, especially for the complex ones, requires superior mastery in two points: perceptual capability of pixel understanding and cognitive ability for semantic understanding. 
\textbf{Firstly}, to achieve precise content perception, \textcolor{blue}{a fine-grained perceptive pixel understanding} of the video movement is necessary. 
Most existing video understanding approaches focus on instance or patch-level analysis \cite{yuan2021tokens,neimark2021video}, lacking the precision for detailed granular control and accurate object-level recognition or tracking, let alone in-depth video comprehension. 
\textbf{Secondly}, profound reasoning demands \textcolor{blue}{cognitive capabilities allowing reasonable explanation and even causal imagination}, i.e., with a reservoir of commonsense knowledge to link video pixels to the factual world.
For example, understanding that jumping from a height can cause fractures, or that colliding with a tanker truck can cause an explosion. 
\textbf{Most importantly}, for humans, video reasoning is not an instantaneous process but follows \textcolor{blue}{a multi-hop procedure from lower level to higher level}.
This often involves first identifying specific targets, like a ``red oil truck'' (cf. Fig. \ref{fig:intro}) in the video frames, then tracking and analyzing its temporal behaviors and interactions with the environment to deduce the scene semantics, and finally, integrating factual commonsense to formulate a cognitively coherent response.

Recently, the community of MLLMs has seen rapid advancement, exhibiting formidable data understanding and reasoning capabilities, among which video MLLMs have been extensively developed, such as Video-LLaMA \cite{abs-2306-02858}, Video-ChatGPT \cite{abs-2306-05424}, and Video-LLaVA \cite{abs-2311-10122}. 
Simultaneously, there is a growing interest in integrating CoT prompting technique \cite{wei2022chain} to augment the reasoning capabilities of LLMs. 
CoT works by intuitively breaking down a complex problem into a chain of simpler and more manageable sub-problems, facilitating a human-like reasoning process. 
While this technique has flourished in language understanding tasks extensively \cite{wang2022self}, unfortunately, a CoT-based reasoning framework specifically tailored for video input with video MLLMs is yet under-explored.

\begin{figure}[!t]
\centering
\includegraphics[width=1\columnwidth]{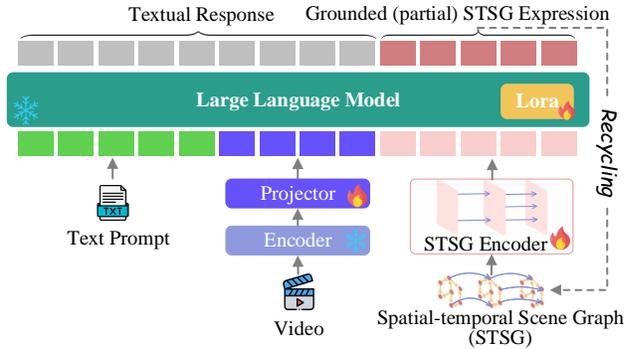}
\caption{
Overview of the MotionEpic video MLLM.
}
  \vspace{-4mm}
\label{fig:framework}
\end{figure}

To this end, this paper is dedicated to devising a solution that enables human-like complex video reasoning. 
We first propose the integration of a STSG representation \cite{ji2020action}, modeling both the input video and its STSG representation, where fine-grained spatial-temporal features are carefully integrated and modeled. 
To implement this, we introduce a novel video LLM, named \textbf{MotionEpic} (cf. Fig. \ref{fig:framework}), which, based on a similar architecture as existing video MLLMs, supports not only video input but also the encoding, understanding and generation of STSGs. 
To enable MotionEpic with fine-grained pixel-level spatial-temporal grounding between videos and STSGs, we also investigate various distinct video-STSG training objects.
STSG annotations are used during the grounding-aware tuning phase, while in the subsequent stage, the system is learned to autonomously parse STSG, and thus supports STSG-free inference and reasoning for downstream tasks.

Building upon MotionEpic, we next design a novel reasoning framework, named Video-of-Thought (\textbf{VoT}), cf. Fig. \ref{fig:VoT}.
Inheriting the key spirit of CoT, VoT breaks down the raw intricate video reasoning problem into a chain of simpler sub-problems, and solves them one by one sequentially.
These sub-questions follow a progression from lower to higher level, i.e., starting with pixel grounding for a precise understanding of target content, and then accurately interpreting corresponding semantic signals. 
\circlenum{1} Given an input video and a question, VoT identifies the possible target(s) involved in the question to observe.
\circlenum{2} The system then grounds the temporal tracklet(s), which serves as supporting evidence/rationale for content perception in subsequent analysis. 
\circlenum{3} Combined with factual commonsense, VoT next interprets the target object's trajectory and its interactions with neighboring scenes to thoroughly understand the action dynamics and semantics.
\circlenum{4} With in-depth understanding of the target actions in the video, we then carefully examine each optional answer with commonsense knowledge, where the final result is output after ranking those candidates.
\circlenum{5} Finally, VoT performs verification for the answer from both pixel grounding perception and commonsense cognition perspectives, ensuring the most factually accurate result.

Our experiments mainly focus on video Question Answering (QA), a representative task reliant on in-depth video reasoning. 
We evaluate our system across 8 complex video QA benchmarks, where it strikingly boosts the current performances in both fine-tuning and zero-shot settings by very clear margins, establishing a series of new states of the arts. 
We further conduct in-depth analyses of MotionEpic's capabilities in video grounding, and probe the video reasoning ability of VoT framework, providing insights into how the framework advances. 
To summarize, this work contributes in multiple aspects: 
\begin{compactitem}
    \item proposing the first video Chain-of-Thought reasoning framework, VoT, which decomposes raw complex problems into a chain of sub-problems, and reasons through multiple steps from low to high levels, enabling not only pixel perceptive recognition but also semantic cognitive understanding of videos.
    
    \item contributing a novel video MLLM, MotionEpic, which supports fine-grained pixel-level spatial-temporal video grounding via STSG encoding and generation. 
    
    \item empirically setting new state-of-the-art (SoTA) performances in a range of video QA benchmarks that require intricate reasoning capability.
\end{compactitem}

\vspace{-2mm}
\section{Related Work}

A key objective in the intelligence community is the understanding of various modalities of data.
Currently, with the advent of LLMs such as ChatGPT \cite{chatgpt}, we have attained unprecedented language reasoning capabilities, on par with the human level. 
This is largely due to the vast repository of commonsense knowledge and semantic understanding capabilities inherent in LLMs, enabling provide plausible causal explanations and even engage in imaginative reasoning. 
Particularly, the integration of the recent trending CoT technology, which deconstructs a problem into its constituent parts and provides rationale at each step, has made the reasoning process more reliable. 
As for image understanding, the rapid development of MLLMs, e.g., LLaVA \cite{abs-2304-08485}, GPT-4V \cite{gpt4}, has also nearly achieved substantial comprehension ability. 
However, unlike language and images, video understanding or reasoning presents a dual challenge of static spatial and temporal dynamics.

Historically, earlier video understanding research efforts predominantly learn neural models over small-size in-domain training datasets \cite{zolfaghari2018eco,lin2019tsm}. 
However, these `small' models are limited to relatively superficial levels of perception, lacking the depth of human-level cognition. 
As a result, previous methods were mostly confined to the shallow understanding of simple videos, such as identifying contents and movements within a video. 
Unlike simple video comprehension, which relies mainly on perceptive abilities, understanding complex videos necessitates deeper cognitive reasoning, such as \emph{explaining why certain actions occur in a video or hypothesizing potential outcomes}. 
Although MLLMs supporting video data have been developed \cite{abs-2305-06355,abs-2306-02858,wu2023nextgpt}, offering greater video understanding capabilities than smaller models, the research into penetrating beyond the perceptual surface of videos to deeply understand the implied semantic content and perform cognitive-level reasoning is still insufficiently explored. 
We observe that current video MLLMs either fail to achieve fine-grained spatial-temporal understanding of videos, or do not fully leverage the rich commonsense knowledge and causal reasoning inherent in LLMs for enhanced cognitive-level comprehension. 
To enable MLLMs with spatiotemporal modeling, we consider employing the dynamic video scene graph representation \cite{ji2020action}.
SGs \cite{JohnsonGF18} are characterized by highly structured graph representations \cite{FeiMatchStruICML22}, which intrinsically depict the underlying semantic implications of the data, and thus have been extensively integrated into a wide range of downstream cross-modal tasks \cite{zhao-etal-2023-generating-visual,wu2024imagine,fei-etal-2023-scene,wu-etal-2023-cross2stra,wu-etal-2023-information}, especially in video modeling \cite{zhao2023constructing,fei2023empowering,fei2024enhancing}.

Meanwhile, recent advancements in CoT technology have made significant strides in enhancing the reasoning capabilities of LLMs \cite{wei2022chain,zhang2022automatic,fei2023reasoning,zheng2024reverse}.
While there are efforts enhancing multimodal reasoning with multimodal CoT \cite{lu2022learn,zhang2023multimodal}, we still note a lack of research specifically focused on integrating CoT into video scenarios to establish a powerful video reasoning framework.
To bridge this gap, this paper takes the initiative and introduces the concept of Video-of-Thought. 
Unlike the original CoT approach that attempts to improve outputs with a simple ``\emph{Let’s Think Step By Step}'' prompt \cite{wei2022chain}, we implement a more genuine thought chain. 
We encourage MLLM to first decompose the original problem into a series of more manageable sub-solutions before the model initiates reasoning, following the human-cognitive procedure from low-level pixel grounding and understanding to high-level cognitive semantic meaning inference, ultimately achieving human-level video understanding and reasoning capabilities.

\section{MotionEpic: Fine-grained Spatial-temporal Grounded Video MLLM}

In this section, we describe the MotionEpic video MLLM, and elaborate on how the STSGs are integrated, as well as the fine-grained spatial-temporal grounding-aware tuning.

\vspace{-2mm}
\subsection{Architecture Briefing}

\vspace{-1mm}
Fig. \ref{fig:framework} presents a schematic overview of MotionEpic, where MotionEpic takes as input three sources: text prompt, video, and STSG representation of video.
We follow the most common practice, and employ the Vicuna-7B (v1.5) \cite{vicuna} as the backbone LLM.
To perceive video input, we adopt the ViT-L/14 encoder \cite{dosovitskiy2020image} and Q-Former projector \cite{0008LSH23}.
We also design MotionEpic to support the STSG signal, where we retrofit the Graph Transformer \cite{dwivedi2020generalization} with recurrent propagation to encode the multi-frame STSG information.

\begin{figure}[!t]
\centering
\includegraphics[width=1\columnwidth]{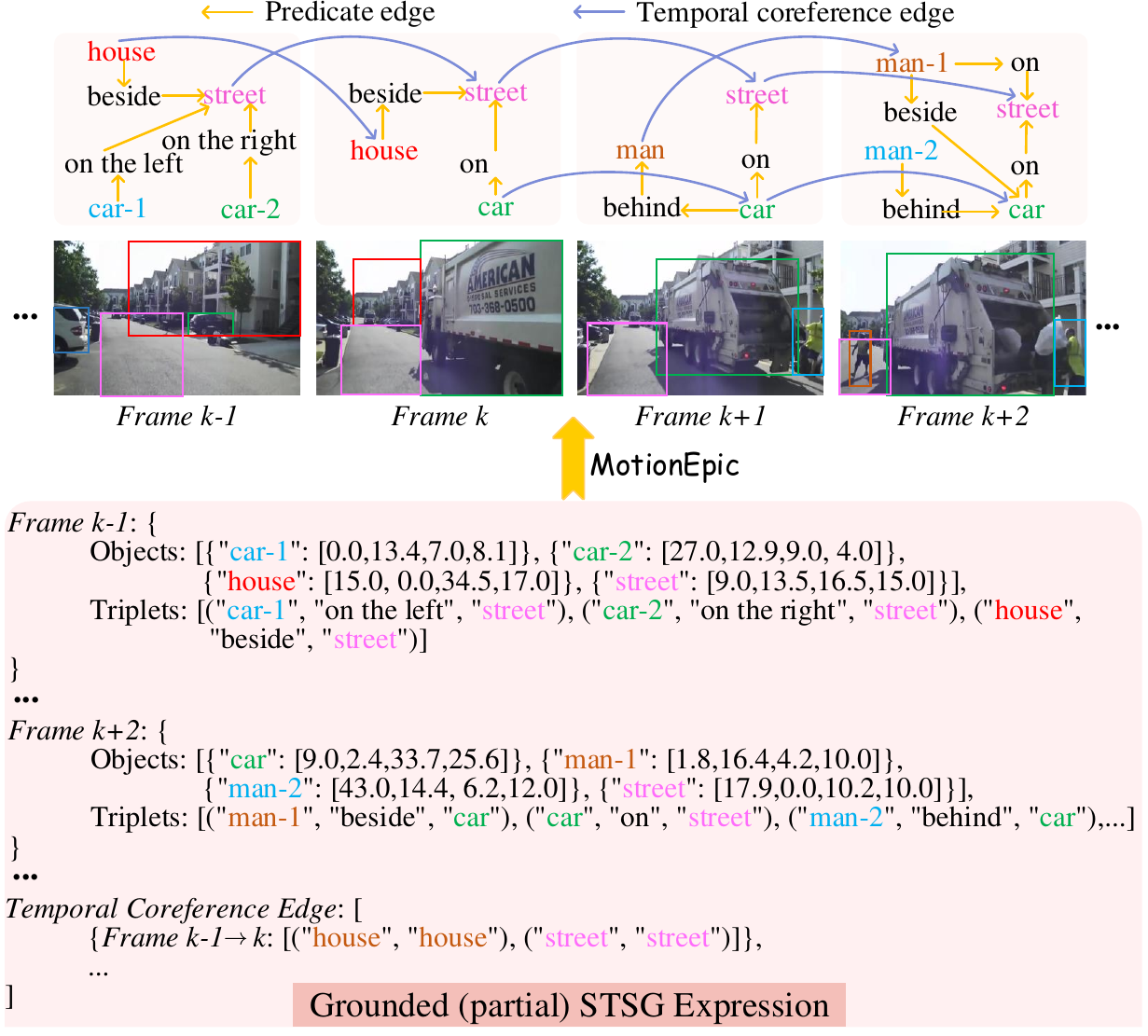}
\caption{
The STSG expression generated by MotionEpic, with its corresponding structural STSG illustration.
}
  \vspace{-5mm}
\label{fig:stsg}
\end{figure}

\vspace{-2mm}
\subsection{Integrating STSG Representation}
\label{Integrating STSG Representation}

\vspace{-1mm}
By definition \cite{ji2020action}, an STSG consists of a sequence of single SGs corresponding to all video frames, with each SG comprising triplets in the video frame, i.e., `\emph{subject}'-`\emph{predicate}'-`\emph{object}', where `\emph{subject}' and `\emph{object}' refer to two visual proposals (RoIs) that are connected with the `\emph{predicate}' relationship.
STSG intuitively depicts the underlying core semantics representations of videos while filtering the less-informative background information, aiding the perceptive understanding of videos \cite{zhao2023constructing}.
Also, such fine-grained structural feature helps effectively model the compositional spatiotemporal semantics.

In our practice, we slightly retrofit the vanilla STSG definition to meet the demand in our reasoning framework.
Since a video has redundant temporal contents across frames, we first evenly sample the frames (with a proper sampling rate), which can effectively reduce computation costs.
We denote each single SG at $k$-th frame as ${G}_k$=$({V}_k;{E}_k)$, where $V_k$ is a list of nodes, i.e., object proposal, and $E_k$ is a list of predicate edges.
For each object proposal $v_{k,i}$ we record the category label $c_i$, the proposal's neural representation $f_i$, and the bounding box (bbox) annotation $b_{i}$=$(x,y,w,h)$, i.e., the 2D coordinate in the image.
Thus, each $v_{k,i}$=$(c_i, f_i, b_{i})_{k}$.
All nodes (i.e., $v_{k,i}$ and $v_{k,j}$) are connected with edges $e_{k,i,j}$.
To enhance the connectivity of STSG, we further create a type of \emph{temporal coreference edges} across each single-frame SG, where the same objects are linked together with time-persistent edges, $e_{k-1\to k}$, mimicking the `tracking' process.

MotionEpic achieves fine-grained spatial-temporal video grounding by simultaneously understanding and generating STSGs.
After full tuning (cf. $\S$\ref{sec:tuning}), MotionEpic can directly output (partial) STSG based on input video (with text prompts), essentially grounding the specific portions of the video content as indicated in the input prompts. 
In Fig. \ref{fig:stsg} we illustrate how the generated STSG expression corresponds to the structural STSG.
Further, the output STSG serving as the rationale will be recycled in the system, i.e., repurposed as the input for the subsequent round.

\vspace{-2mm}
\subsection{Fine-grained Video-Scene Grounding-aware Tuning}
\label{sec:tuning}

\vspace{-1mm}
Intuitively, we expect our system to perform video reasoning for downstream tasks without relying on any external STSG annotations, i.e., STSG-free inference.
This requires an accurate spatial-temporal grounding between videos and STSGs.
To this end, we carry out tuning for MotionEpic such that it is learned to autonomously parse STSG according to input instructions.
The grounding-aware tuning is performed based on video-STSG pairs.
We design various training objectives, which can be further divided into coarse-grained and fine-grained levels:

\vspace{-1mm}
\textbf{1) Enhancing coarse-grained correspondence}:
\vspace{-2mm}
\begin{compactitem}
    \item $\mathcal{L}_1$: predicting if the overall input video and STSG are paired.

    \item $\mathcal{L}_2$: given a video, generating the whole STSG (expression) of the video.
        
\end{compactitem}

\vspace{-1mm}
\textbf{2) Enhancing fine-grained correspondence}:
\vspace{-2mm}
\begin{compactitem}
    \item $\mathcal{L}_3$: given a video and action description(s), outputting the corresponding object tracklet(s), i.e., a partial STSG.
    
    \item $\mathcal{L}_4$: given a video and key object(s), describing the corresponding temporal action(s) in textual response, and outputting the corresponding object tracklet(s).

    \item $\mathcal{L}_5$: given a video and a bbox of a certain frame's object, outputting the object label, as well as the corresponding tracklet.
        
\end{compactitem}

For each learning objective, we wrap up the inputs with instruction-tuning \cite{abs-2304-08485} style question-answer pairs, being consistent with the following downstream inference.
Overall, except for the STSG encoder and video projector, the video encoder and the backbone LLM are kept frozen throughout all the learning stages.
To tune the LLM, we leverage LoRA \citep{HuSWALWWC22} to enable a small subset of parameters to be updated.

Before conducting the above grounding-level tuning, we perform conventional video pre-training on Webvid, which serves as the important warming starting for the following video understanding tuning,
Despite aligning the encoding modules with LLM, there remains a gap towards the goal of enabling the overall system to faithfully follow and understand users' instructions and generate the desired outputs. 
To address this, further instruction tuning is necessary.
After the grounding-level tuning, we utilized existing video instruction tuning data for instruction tuning of the model, which includes the dataset from VideoChat \citep{abs-2305-06355} and Video-ChatGPT \citep{abs-2306-05424}

\section{Video-of-Thought Reasoning Framework}
\label{sec:vot}

\begin{figure}[!t]
\centering
\includegraphics[width=1\columnwidth]{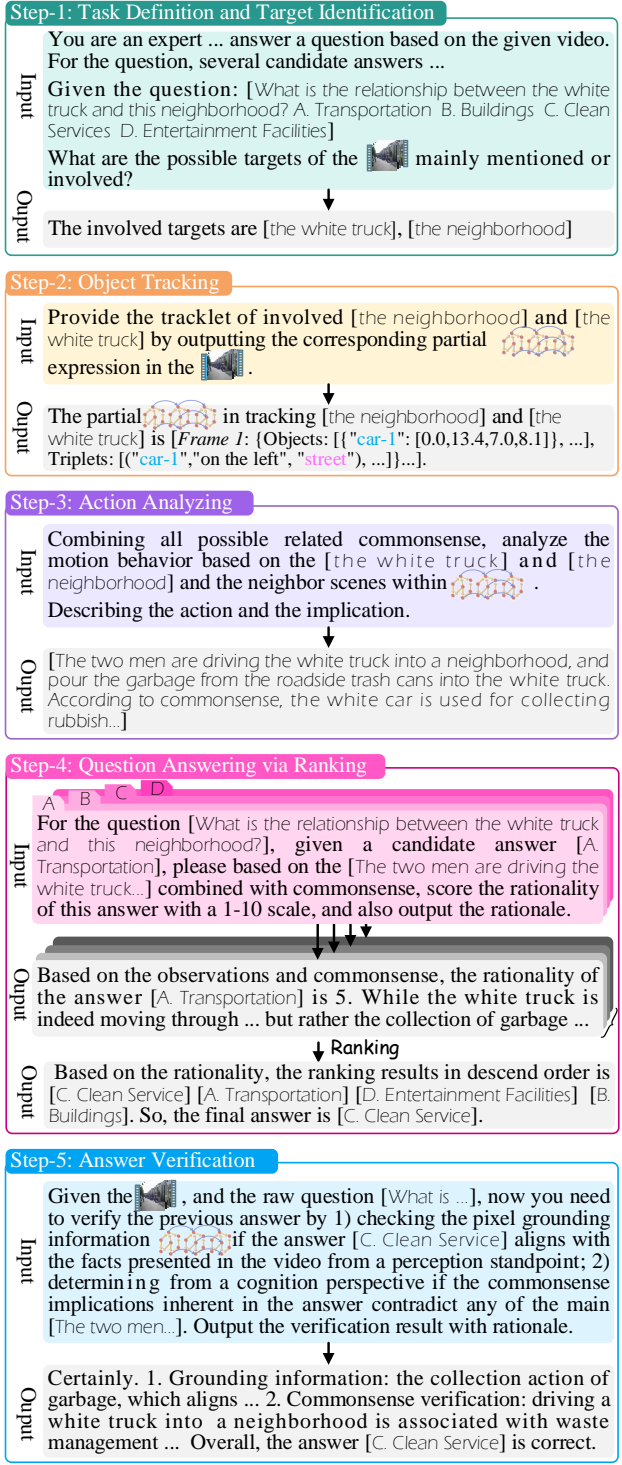}
\caption{
An illustrative view of VoT framework.
The complete I/O and prompts are detailed in Appendix.
}
\vspace{-3mm}
\label{fig:VoT}
\end{figure}

Based on MotionEpic, we now perform video reasoning with VoT.
Different from the vanilla CoT with one straightforward prompt, i.e., ``\emph{Let's think step by step}'', VoT divides the raw problem into much smaller and finer-grained sub-problems.
We consider an exact paradigm of task decomposition, which encompasses five chained steps, following a process from low-level perceptive pixel grounding to high-level cognitive semantic comprehension.
In Fig. \ref{fig:VoT} we illustrate the overall VoT framework.

\vspace{-3mm}
\paragraph{$\blacktriangleright$ Step-1: Task Definition and Target Identification\\}
First, MotionEpic is fed with the raw video along with the text prompt of the task definition, format, and raw question, all of which serve as the background foundation information of the reasoning.
As the initial phase, we expect to identify the target within the video that requires analysis, which is a crucial prerequisite for determining the subsequent in-depth reasoning. 
It is noteworthy that sometimes the question may explicitly include targets visible in the video, or implicitly involve related targets. 
Therefore, we proceed to prompt the model, to infer from the question what the target object(s) involved or related to in the video might be:
\begin{mybox}\texttt
Given the question [\texttt{Question}], what are the possible targets of the \videologo mainly mentioned or involved?
\end{mybox}
After this step, all the possible [\texttt{Target}] involved in the question will be confirmed.

\vspace{-3mm}
\paragraph{$\blacktriangleright$ Step-2: Object Tracking\\}
In the second step, we aim to further ground the object's full spatial-temporal characteristics, i.e., to track the target's trajectory.
We note that grounding the targets' temporal tracking is pivotal for pursuing fine-grained video understanding, as only accurately perceiving the behaviors in the video can ensure that the subsequent cognitive-level understanding is meaningful. 
In this work, we leverage the STSG for the temporal grounding, rather than directly tracking the original video frames. 
Such semantic representation carried by STSG is highly concise, ensuring that the tracking of the video's target is more accurate and reliable. 
Also notably, object tracking and pixel grounding based on the STSG can effectively mitigate the hallucination issues \cite{zhang2023siren} inherent in existing MLLMs.

Having performed grounding-aware tuning, MotionEpic develops the full capability to ground from object to (partial) STSG. 
Therefore, we directly prompt the model with:
\begin{mybox}\texttt
Provide the tracklet of involved [\texttt{Target}] by outputting the corresponding partial \stsglogo expression.
\end{mybox}
The yielded grounded [\texttt{Target Tracklet}] of STSG will serve as low-level evidence (i.e., supporting rationale) for the next step of behavior analysis.

\vspace{-3mm}
\paragraph{$\blacktriangleright$ Step-3: Action Analyzing\\}
In this step, VoT further analyze the corresponding actions and behaviors by integrating the target tracking in STSG. 
For an accurate understanding of the target object's motion, merely observing the target itself is insufficient. 
This process should also reference the higher-order neighbor nodes within the STSG representation, interacting targets with their neighboring scenes.
On the other hand, directly inferring actions from video pixels alone is still inadequate, as interpretations based solely on pixel information often remain superficial. 
Therefore, we further prompt the model to consider more potentially relevant commonsense knowledge, allowing the model to connect video pixel observations with the factual world, achieving a more in-depth understanding of the video. 
Given that MLLMs possess the necessary repository of commonsense knowledge via extensive pre-training, all that is required is to properly prompt the model:
\begin{mybox}\texttt
Combining all possible related commonsense, analyze the motion behavior based on the [\texttt{Target Tracklet}] and the neighbor scenes within \stsglogo.
Describing the action observations and implications.
\end{mybox}
This step yields the target action's [\texttt{Observation and Implication}].

\vspace{-3mm}
\paragraph{$\blacktriangleright$ Step-4: Question Answering via Ranking\\}
Having established an in-depth understanding of the target actions in the video, we can now consider answering the original question. 
We contemplate a multiple-choice QA format, where multiple candidate answers are provided.\footnote{
Note that for open-ended QA, we consider prompting the model to output multiple distinct optional answers,
such that we unify different types of QA formats into a multi-choice format. 
For the open-ended QA format, we detail processing and prompt methods in Appendix.
}
Inspired by the human pattern of answering multi-choice questions, we also consider a ranking mechanism to determine the final answer.
Specifically, for each candidate answer, we prompt the model to score its likelihood (from 1 to 10) in conjunction with commonsense knowledge, and provide a corresponding rationale: 
\begin{mybox}\texttt
For the question [\texttt{Question}], given a candidate answer [\texttt{Answer}], please based on the action's [\texttt{Observation and Implication}] combined with commonsense, score the rationality of this answer with a 1-10 scale, and also output the rationale.
\end{mybox}
We then rank the scores of all options and select the most optimal answer [\texttt{Answer}].

\vspace{-3mm}
\paragraph{$\blacktriangleright$ Step-5: Answer Verification\\}
Given that complex video task often involves intricate questions and answers, and the entire reasoning process encompasses lengthy chained steps, it is essential to verify the answer provided in the previous step.
Our basic idea to verification is that, assuming that answer A is correct, we retrospectively evaluate whether the answer results in contradictions with the input question and video in two aspects:
1) First, check the pixel grounding information if it aligns with the facts presented in the video from a perception standpoint.
2) On the other hand, prompt the model again from a cognition perspective to determine if the commonsense implications inherent in the answer contradict any of the main observations inferred in the 3-$rd$ reasoning step.
\begin{mybox}\texttt
Given the \videologo, and the raw question [\texttt{Question}], now you need to verify the previous answer by\\
\hspace*{10pt}  1) checking the pixel grounding information if the answer [\texttt{Answer}] aligns with the facts presented in the video from a perception standpoint; \\
\hspace*{10pt}  2) determining from a cognition perspective if the commonsense implications inherent in the answer contradict any of the main [\texttt{Observations}] inferred in the 3-$rd$ reasoning step.\\
Output the verification result with rationale.
\end{mybox}
If any inconsistencies are found in perception and cognition perspectives, we record the corresponding rationale, and re-execute the 4-$th$ step to reselect the answer.
This approach ensures that the final outcome is the most factually accurate.

\section{Experiments}

\subsection{Settings}

\paragraph{Task and Data.}
While in theory all video understanding tasks could benefit from our reasoning framework, we mainly focus on the most representative task, video QA.
For fine-tuning setting, we adopt 6 benchmarks characterizing complex video QA where advanced video abilities, e.g., explanation, causality, foresight and imagination are required: VLEP \cite{lei2020more}, STAR \cite{wu2021star}, IntentQA \cite{li2023intentqa}, Social-IQ \cite{zadeh2019social}, Causal-VidQA \cite{li2022representation} and NExT-QA \cite{XiaoSYC21}.
For zero-shot setting, we further consider using MSR-VTT \cite{XuMYR16} and ActivityNet \cite{HeilbronEGN15} datasets.
All datasets come with their own splitting, and we follow the prior practice without modification.

\vspace{-3mm}
\paragraph{Grounding-aware Tuning Corpus.}
To construct the video-STSG pairs, we leverage the Action Genome data \cite{ji2020action}, which contains 10K high-quality manual annotated STSGs of videos.
To enrich the data amount, we also use part of WebVid-10M videos \cite{bain2021frozen}, where we select 350K videos with clear actions, and parse the STSGs via SoTA parser \cite{li2022dynamic}.

\vspace{-3mm}
\paragraph{Implementations.}
MotionEpic uses the Vicuna-7B (v1.5)\footnote{\url{https://huggingface.co/lmsys/vicuna-7b-v1.5}} as the backbone LLM.
We adopt the ViT-L/14\footnote{\url{https://huggingface.co/sentence-transformers/clip-ViT-L-14}} as the video encoder, and use the Q-Former\footnote{\url{https://huggingface.co/spaces/Salesforce/BLIP2}} as the projector.
All the modules take the default configurations without much modification.
For our recurrent Graph Transformer encoding STSGs, we take a 6-layer architecture with 768-d hidden sizes.
The object neural representation $f_i$ is obtained via CLIP, which will be used as node embedding initiation.
The text tokenizer is sourced from LLaMA, with approximately 32,000 classes.
For each video, we uniformly sample certain frames with a sampling rate of 8 fps for fine-grained reasoning. 
We note that too large sampling rate introduces noises (i.e., redundant frames) and huge computation costs, while too small one will cause important information loss.
Here we use the 8 fps, as in our preliminary study we verified that it helps achieve the best trade-off.
For the fine-tuning setting of end tasks, we will tune the MotionEpic based on the training set using the setting as prior baselines, i.e., data split and evaluation methods.
For the zero-shot setting, we will directly perform video QA without using the in-domain training set.
All trainings are conducted on 16 NVIDIA A100 GPUs.

\vspace{-3mm}
\paragraph{Baselines and Evaluations.}
The evaluations are compared with recent SoTA baselines of these complex video QA datasets, including 
InternVideo \cite{wang2022internvideo},
LLaMA-VQA \cite{ko2023large},
VLAP \cite{wang2023vlap},
SeViLA \cite{yu2023self},
TranSTR \cite{li2023discovering} and
HiTeA \cite{ye2023hitea}.
The results are faithfully copied from their papers.
Also we reimplement current video MLLMs, including
VideoChat2 \cite{li2023mvbench},
Video-LLaMA \cite{abs-2306-02858},
Video-ChatGPT \cite{abs-2306-05424},
VideoChat \cite{li2023videochat} and
Video-LLaVA \cite{abs-2311-10122}.
For fairness, we compare MotionEpic with these video MLLMs in a vanilla CoT setting.
Further, we also implement the Video-LLaVA by integrating the STSG features.
We adopt accuracy as the main metric of the QA task performance, following the prior practice.

\begin{table}[!t]
\centering
\fontsize{8}{11}\selectfont
\setlength{\tabcolsep}{0.6mm}
\captionof{table}{
\label{tab:main-1}
Results on four VideoQA datasets.
STAR data includes four subsets: Interaction (Int.), Sequence (Seq.), Prediction (Pre.), Feasibility (Fea.).
The best scores of baselines are \underline{underlined}, and the new best results are \textbf{bold}.
}
\vspace{-1mm}
\begin{tabular}{lccccccccc}
\hline
\multicolumn{1}{c}{\multirow{2}{*}{\bf Model}} & \multicolumn{1}{c}{\multirow{2}{*}{\bf VLEP}}& \multicolumn{4}{c}{\multirow{1}{*}{\bf STAR}}& \multicolumn{1}{c}{\multirow{2}{*}{\bf IntentQA}}& \multicolumn{2}{c}{\multirow{1}{*}{\bf Social-IQ}} \\
\cmidrule(r){3-6}\cmidrule(r){8-9}
& &  Int.& Seq.&  Pre.&  Fea.& & 2-Way& 4-Way \\
\hline
\multicolumn{9}{l}{$\bullet$ \bf SoTA baselines}\\
InternVideo& 	63.9& 	62.7&  	65.6&  	54.9&  	51.9& 	-& 	-& 	-\\
LLaMA-VQA& 	\underline{71.0}& 	66.2&  	67.9 & 	57.2&  	52.7& 	-& 	-& 	-\\
VLAP& 	69.6& 	\underline{70.0} & 	\underline{70.4} & 	\underline{65.9}& 	\underline{62.2}& 	-& 	-& 	-\\
SeViLA& 	68.9 & 	63.7&  	70.4&  	63.1&  	62.4& 	-& 	-& 	-\\
VideoChat& 	62.0& 	63.2& 	66.8& 	54.1& 	49.6& 	59.3& 	67.7& 	37.8\\
Video-LLaVA& 	65.8& 	64.3& 	67.0& 	56.5& 	50.1& 	62.5& 	68.9& 	39.2\\

\hline
\multicolumn{9}{l}{$\bullet$ \bf CoT}\\
Video-LLaVA&	65.7&	65.0&	67.7&	57.8&	52.0&	63.2&	69.5&	40.4\\
Video-LLaVA\tiny{+STSG}&	67.0&	65.9&	68.9&	58.7&	53.7&	64.9	&70.4&	41.7\\
MotionEpic&	68.2&	66.8 &	69.6& 	60.6 &	57.4&	\underline{66.1}&	\underline{71.7}&	\underline{43.0}\\

\hline
\multicolumn{9}{l}{$\bullet$ \bf VoT}\\
\rowcolor{bluei} MotionEpic&		\bf 73.4&		\bf 71.5&		\bf 72.6&		\bf 66.6&		\bf 62.7&		\bf 70.8&		\bf 72.8&		\bf 45.0\\

\hline
\end{tabular}
\vspace{-3mm}
\end{table}

\begin{table}[!t]
\centering
\fontsize{8}{11}\selectfont
\setlength{\tabcolsep}{0.8mm}
\captionof{table}{
\label{tab:main-2}
Results on Causal-VidQA data.
D: Description, E: Explanation, P: Prediction, C: Counterfactual. 
}
\vspace{-1mm}
\begin{tabular}{lcccccccc}
\hline
\multicolumn{1}{c}{\multirow{2}{*}{\bf Model}} & \multicolumn{1}{c}{\multirow{2}{*}{\bf Acc@D}}& \multicolumn{1}{c}{\multirow{2}{*}{\bf Acc@E}}& \multicolumn{3}{c}{\multirow{1}{*}{\bf Acc@P}}& \multicolumn{3}{c}{\multirow{1}{*}{\bf Acc@C}} \\
\cmidrule(r){4-6}\cmidrule(r){7-9}
& &  & A& 	R& 	AR& 	A& 	R& 	AR \\
\hline
\multicolumn{9}{l}{$\bullet$ \bf SoTA baselines}\\
TranSTR&	73.6&	75.8&	65.1&	65.0&	48.9&	68.6&	65.3&	50.3\\
Video-LLaMA&	69.2&	71.0&	63.6&	62.4&	44.4&	65.4&	60.1&	45.0\\
VideoChat&	72.9&	73.9&	65.2&	63.1&	45.9&	66.0&	62.7&	45.8\\
Video-ChatGPT&	73.1&	75.1&	66.0&	63.9&	46.0&	67.8& 63.6&	50.0\\
Video-LLaVA&	{73.7}&	{74.4}&	{67.6}&	{65.4}&	{47.7}&	{68.0}&	{64.9}&	{51.5}\\

\hline
\multicolumn{9}{l}{$\bullet$ \bf CoT}\\
Video-LLaVA&	74.2&	74.8&	68.0&	65.7&	48.1&	70.3	&65.7&	52.9\\
Video-LLaVA\tiny{+STSG}&	75.7&	75.9&	68.9&	67.2&	50.0&	70.7&	67.2&	53.6\\
MotionEpic&	\underline{78.5}&	\underline{77.2}&	\underline{70.1}&	\underline{70.8}&	\underline{52.4}&	\underline{71.2}&	\underline{69.1}&	\underline{55.0}\\

\hline
\multicolumn{9}{l}{$\bullet$ \bf VoT}\\
\rowcolor{bluei} MotionEpic	&\bf 81.2	&\bf 83.0	&\bf 74.3	&\bf 73.7	&\bf 54.7	&\bf 74.5	&\bf 73.8	&\bf 58.6\\

\hline
\end{tabular}
\end{table}

\vspace{-1mm}
\subsection{Main Performance on Video QA Reasoning}

\vspace{-1mm}
In Table \ref{tab:main-1}, \ref{tab:main-2} and \ref{tab:main-3} we present the main results of different systems.
Overall, our MotionEpic under the VoT reasoning framework has boosted all the SoTA baselines by very large margins consistently. 
Beyond enhanced performance, we further gain some key observations.
First, by observing Video-LLaVA without/with CoT prompting, we see that the improvement from CoT for video reasoning could be quite limited.
Further, by comparing Video-LLaVA without/with STSG integration, we notice that the structural fine-grained STSG features play a positive role in understanding videos.
Third, by comparing Video-LLaVA+STSG with our MotionEpic under the same CoT, it is clear that the implicit integration of the scene graph features is quite superior to the explicit integration.
Also, even our MotionEpic with vanilla CoT we beat the SoTA methods on certain datasets.
Lastly, observing the MotionEpic under CoT and our proposed VoT, we see there are huge performance gaps in between consistently on all reasoning scenarios and tasks, indicating the great potential of our proposed video reasoning framework.

\begin{table}[!t]
\centering
\fontsize{8}{11}\selectfont
\setlength{\tabcolsep}{1.7mm}
\captionof{table}{
\label{tab:main-3}
Results on  NExT-QA data.
}
\vspace{-1mm}
\begin{tabular}{lcccc}
\hline
\bf Model & \bf Acc@All & \bf 	Acc@C & \bf  	Acc@T & \bf  	Acc@D\\
\hline
\multicolumn{5}{l}{$\bullet$ \bf SoTA baselines}\\
InternVideo & 	63.2 & 	62.5  & 	58.5 & 	75.8\\
HiTeA & 	63.1 & 	62.4  & 	58.3 & 	75.6\\
LLaMA-VQA & 	72.0 & 	72.7  & 	69.2  & 	75.8\\
SeViLA & 	73.8 & 	73.8  & 	67.0 &  	81.8\\
VLAP & 	\underline{75.5} & 	\underline{74.9} & 	\underline{72.3} & 	\underline{82.1}\\
Video-LLaMA & 	60.6 & 	59.2 & 	57.4 & 	72.3\\
VideoChat & 	61.8 & 	63.5 & 	61.5 & 	74.6\\
Video-ChatGPT & 	64.4 & 	66.9 & 	64.1 & 	75.7\\
Video-LLaVA & 	66.3 & 	67.7 & 	63.8 & 	75.9\\

\hline
\multicolumn{5}{l}{$\bullet$ \bf CoT}\\
Video-LLaVA &	67.7 &	69.0 &	65.9 &	76.5\\
Video-LLaVA\tiny{+STSG} &	68.0 &	71.6 &	67.6 &	78.9\\
MotionEpic &	72.2 &	73.4 &	69.1 &	80.7\\

\hline
\multicolumn{5}{l}{$\bullet$ \bf VoT}\\
\rowcolor{bluei} MotionEpic	&\bf 76.0 &\bf 75.8  &\bf 74.6  &\bf 83.3 \\

\hline
\end{tabular}
\vspace{-3mm}
\end{table}

\begin{table}[!t]
\centering
\fontsize{8}{11}\selectfont
\setlength{\tabcolsep}{0.6mm}
\captionof{table}{
\label{tab:few-shot}
Zero-shot Video QA results.
Verify-G/C: verification in terms of \underline{G}rounding and \underline{C}ommonsense perspectives.
}
\vspace{-1mm}
\begin{tabular}{lccccc}
\hline
\bf Model & \bf MSR-VTT & \bf 	ActivityNet & \bf 	NExT-QA & \bf 	STAR & \bf \em	AVG.\\

\hline
\multicolumn{5}{l}{$\bullet$ \bf Zero-shot SoTA baselines}\\
InternVideo &	- &	- &	49.1 &	41.6 &	- \\
Video-LLaMA &	49.6 &	21.4 &	43.5 &	36.4 &	37.7 \\
VideoChat &	52.0 &	26.5 &	52.8 &	45.0 &	44.1 \\
Video-ChatGPT &	54.3 &	35.2 &	53.0 &	48.7 &	47.8 \\
Video-LLaVA &	59.2 &	45.3 &	57.3 &	50.6 &	53.1 \\
VideoChat2 &	54.1 &	49.1 &	61.7 &	\underline{59.0} &	56.0 \\

\hline
\multicolumn{5}{l}{$\bullet$ \bf CoT}\\
Video-LLaVA &	60.0 &	46.9 &	59.5 &	52.0 &	54.6\\
Video-LLaVA\tiny{+STSG} &	61.5 &	48.4 &	60.6 &	52.7 &	55.8 \\
MotionEpic &	\underline{63.1} &	\underline{50.0} &	\underline{61.9} &	{56.5} &	\underline{57.8}\\

\hline
\multicolumn{5}{l}{$\bullet$ \bf VoT}\\
\rowcolor{bluei} MotionEpic &	\bf 66.2 &	\bf 54.6 &	\bf 66.5 &	\bf 61.7 &	\bf 62.3 \\

\cdashline{1-6}
\quad w/o Verify-G &	63.6 &	 51.4  &62.0 & 59.1 & 59.0 \\

\quad w/o Verify-C &	65.1&	 53.4	 & 62.8 &	 58.8 &	60.1\\

\hline
\end{tabular}
\end{table}

\vspace{-2mm}
\subsection{Zero-shot Performance}

\vspace{-1mm}
We then examine the performance in zero-shot setting.
Table \ref{tab:few-shot} presents the comparisons.
In general, we can notice that CoT exhibits stronger improvements than direct prompting methods under the zero-shot scenario, compared with the scenario of the above fine-tuning.
Notedly, the improvements by VoT over the CoT become larger and clearer under the zero-shot setting.
The enhancements on these two complex video QA tasks are clearer than those on the comparatively simpler tasks, i.e., MSR-VTT and ActivityNet.
This is largely because the latter datasets more tend to perceptive understanding (e.g., describing \emph{what's in video}), rather than cognitive understanding (e.g., explanation, foresight or imagination).
Further, we cancel the verification mechanism (at 6-$th$) of either the pixel grounding perspective or the commonsense perspective.
We see that on MSR-VTT and ActivityNet, the perceptive-level pixel grounding verification is more crucial than the commonsense cognitive verification.
For those complex videos, both types of verifications are pivotal.

\begin{figure}[!t]
\centering
\includegraphics[width=0.98\columnwidth]{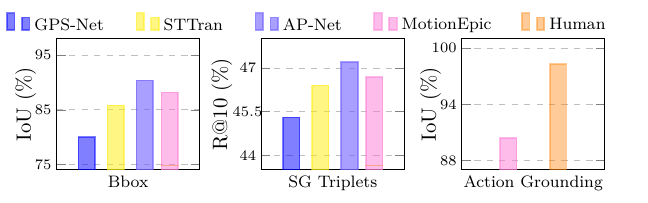}
  \vspace{-1mm}
\caption{
MotionEpic performance on object grounding, scene graph triplet classification, and action grounding.
}
\label{fig:video-grounding}
\vspace{-4mm}
\end{figure}

\vspace{-1mm}

\subsection{Analyses on MotionEpic Video MLLM}

\vspace{-1mm}
\paragraph{Probing Video Grounding Ability.}
To evaluate how well our MotionEpic is capable of video grounding, we here perform the probing test.
Specifically, we evaluate the performance of MotionEpic on STSG parsing on the Action Genome test set, by comparing with SoTA DSG parsers: GPS-Net \cite{lin2020gps}, STTran \cite{cong2021spatial} and AP-Net \cite{li2022dynamic}.
We measure three aspects: 
1) the object grounding (bbox detection),
2) SG triplet classification (categories of entities, and relation predicates between entities),
and 3) temporal action grounding (the start and end times of actions).
Fig. \ref{fig:video-grounding} illustrates the results, where we see that MotionEpic achieves very competitive performance on par with SoTA parser, even with human-level performance.
This reveals that MotionEpic shows reliable capability in providing video grounding information to support the subsequent in-depth video reasoning.

\begin{figure}[!t]
\centering
\includegraphics[width=1\columnwidth]{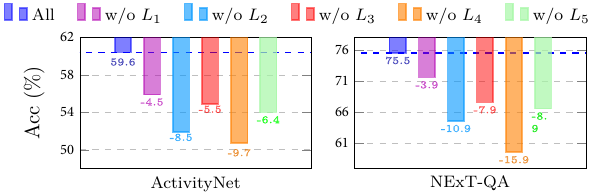}
\caption{
Performance (zero-shot) drop of MotionEpic after ablating different grounding-aware tuning items.
}
\label{fig:grounding-tuning-objectives}
\vspace{-3mm}
\end{figure}

\begin{minipage}[t]{1\columnwidth}
\begin{minipage}[t]{1\columnwidth}
\centering
\fontsize{8}{10}\selectfont
\setlength{\tabcolsep}{1.5mm}
\begin{tabular}{lcccc}
\hline
\multicolumn{1}{c}{\multirow{2}{*}{Data}} &  \multicolumn{2}{c}{CoT} & \multicolumn{1}{c}{VoT} & \multicolumn{1}{c}{\multirow{2}{*}{Human}} \\ 
\cmidrule(r){2-3} \cmidrule(r){4-4}
& Video-LLaVA &MotionEpic & MotionEpic &  \\
\hline 
Causal-VidQA	& 32.4&56.8 &74.3 &80.6  \\
Social-IQ 	& 22.3& 40.1& 61.4&  72.7\\
\hline
\end{tabular}
\end{minipage}
\begin{minipage}[t]{1\columnwidth}
\centering
\includegraphics[width=0.99\columnwidth]{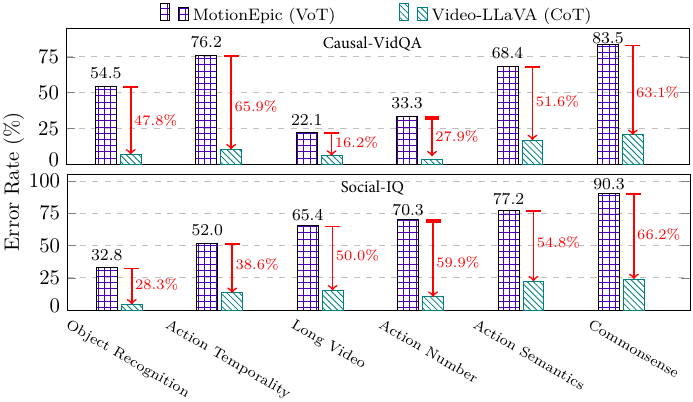}
  \vspace{-3mm}
\captionof{figure}{
\label{fig:error}
\textbf{Above Table}: human evaluation of video QA.
\textbf{Below Figure}: error rate under various specific categories.
}
\end{minipage}
\end{minipage}

\paragraph{Influence of Various Grounding-aware Tuning Strategies.}
We further study the impacts/contributions of different grounding-aware tuning objectives introduced in $\S$\ref{sec:tuning}.
We design five groups of ablations where each tuning goes without one item, and the resulting model performs zero-shot end task.
The results on two datasets are shown in Fig. \ref{fig:grounding-tuning-objectives}, where different items come with varied impacts, indicating the importance of video-STSG grounding fine-tuning.
Notably, the lack of $\mathcal{L}_2$ and $\mathcal{L}_4$ result in the greatest degradation.
This is intuitive, as these two objectives are directly associated with the subsequent reasoning process, i.e., understanding STSG from video, and generating (partial) STSG given objects.

\begin{figure*}[!t]
\centering
\includegraphics[width=1\textwidth]{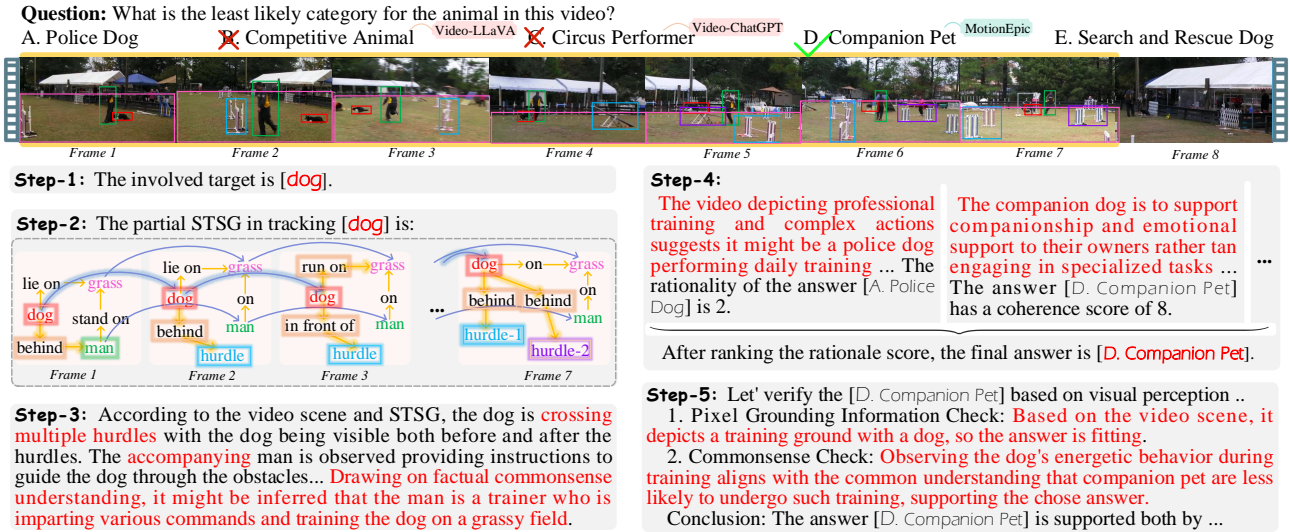}
\caption{
Visualization of qualitative example showcasing how our VoT framework achieves successful video reasoning.
}
\label{fig:demos}
\vspace{-4mm}
\end{figure*}

\subsection{Analyses on VoT Reasoning Framework}

\vspace{-1mm}
\paragraph{Reasoning Ability Breaking-down.}
Previously, we validated the overall stronger performance of the VoT reasoning system through extensive experimentation. 
Here, we aim to provide a more in-depth analysis of VoT. 
First, we select 200 hard instances each from the Causal-VidQA and Social-IQ test sets, and then compare the performance of Video-LLaVA and MotionEpic under CoT and VoT frameworks, respectively. 
Also we conduct human evaluation on this subset to gauge its difficulty level. 
The results, in the above table of Fig. \ref{fig:error}, show that MotionEpic with VoT reasoning framework achieves quite exceptional results, comparable even to human performance.
We further summarize the error cases and analyze differences in the 6 most frequent categories of errors.
As seen in the below part of the figure, MotionEpic (with VoT) significantly reduces the error rate of Video-LLaVA (with CoT), especially in terms of action semantics and commonsense understanding.

\vspace{-3mm}
\paragraph{Video Reasoning Visualization.}
Finally, we present a case study to aid an intuitive understanding of the superiority of our system.
We randomly select an instance where our model gives the correct answer.
As shown in Fig. \ref{fig:demos}, the video displays a complex scene, and the given question is abstract and complex, not directly answerable through the mere perception of the video itself. 
However, our MotionEpic provides the correct answer, while the other two baselines err. 
At the content perception level, VoT ensures accurate and robust understanding through STSG-based video grounding, preventing hallucination, i.e., it correctly interprets that the animal is a dog, then infers from commonsense that the scene involves a trainer training the dog. 
Then, at the cognitive level, it analyzes each option to determine the best answer. 
Through further verification, the result aligns with both the video content and factual commonsense understanding. 
Overall, the entire reasoning greatly improves the accuracy at each step through problem decomposition, while ensuring an explainable process decision rationale.
In the Appendix, we show more qualitative visualizations of examples.

\vspace{-1mm}
\section{Conclusion}

\vspace{-1mm}

In this work, we for the first time introduce an innovative solution for complex video reasoning, the Video-of-Thought (VoT) framework.
To accomplish the reasoning framework, also a novel video MLLM, MotionEpic, is proposed.
MotionEpic achieves fine-grained pixel-level spatial-temporal video grounding by adeptly integrating video STSG representation.
With MotionEpic, the VoT framework resolves the intricate video task by skillfully dissecting it into manageable sub-problems, tackling them sequentially from low-level pixel perception to advanced cognitive interpretation. 
Our experiments across various complex video QA benchmarks have not only proven the efficacy of our approach but have also boosted the existing state-of-the-art standards. 
Overall, this work marks a substantial contribution to the video modeling community, paving the way for more nuanced, human-level analysis in the relevant community.

\vspace{-2mm}
\section*{Acknowledgements}
\vspace{-1mm}
This research is supported by the National Research Foundation Singapore under its AI Singapore Programme (Award Number: AISG-GC-2019-001-2A).
This work is supported by CCF-Baidu Open Fund, and the National Natural Science Foundation of China (NSFC) Grant (No. 62336008).

\vspace{-2mm}
\section*{Statement of Potential Broader Impact}

\vspace{-1mm}
This paper aims to construct a robust, human-level video understanding and reasoning framework. 
The system must be built upon existing LLM to realize its full potential.
Potential implications include substantial energy consumption during LLM system training, leading to environmental degradation, and the necessity for more extensive data corpora for training.
Moreover, due to the powerful video reasoning and comprehension capabilities, there exists the potential for malicious actors to exploit this framework for nefarious intents, posing a societal threat. 
Consequently, the release of this framework necessitates the establishment of specific licensing mechanisms to ensure responsible deployment and mitigate potential misuse.

\newpage

\bibliography{ref}
\bibliographystyle{icml2024}

\newpage
\appendix
\onecolumn

\section{More Configuration Details}

\subsection{Detailed Prompt Construction and System I/O}
\label{Detailed Prompt Construction}
Here, we provide detailed prompts as well as their inputs and outputs, for each step of the VoT reasoning framework.

\paragraph{$\blacktriangleright$ Step-1:}
If the raw question is a multi-choice question, the prompt for Step-1 should be: 
\begin{tcolorbox}[title=Step-1: Task Definition and Target Identification for Multi-choice Question, colback=gray!5,colframe=nmgray!75!black]
$\blacktriangleright$ \textbf{Input:}\\
$<$Task Definition$>$\\ 
Now you are an expert in analyzing video data, and you should answer a question based on the given video.\\
For the question, several candidate answers are provided, where you need to choose [the most suitable option | all possible correct option(s)].\\
$<$/Task Definition$>$\\

$<$Input Video$>$ \videologo $<$/Input Video$>$\\ 

$<$Question$>$\\ 
Given the question: [\texttt{What is the relationship between the white truck and this neighborhood? A. Transportation  B. Buildings  C. Clean Services  D. Entertainment Facilities}], what are the possible targets of the [\texttt{Video}] mainly mentioned or involved?\\
$<$/Question$>$\\

\vspace{2mm}
$\blacktriangleright$ \textbf{Output:}\\
The involved targets are [\texttt{the white truck}], [\texttt{the neighborhood}]
\end{tcolorbox}

Otherwise for the open-ended format, the prompt will be:
\vspace{-4mm}
\begin{tcolorbox}[title=Step-1: Task Definition and Target Identification for Open-ended Question, colback=gray!5,colframe=nmgray!75!black]
$\blacktriangleright$ \textbf{Input:}\\

$<$Task Definition$>$\\ 
Now you are an expert in analyzing video data, and you should answer a question based on the given video.\\
For the question, you should answer in an open-ended format.\\
$<$/Task Definition$>$\\

$<$Video$>$ \videologo $<$/Video$>$\\ 

$<$Question$>$\\ 
Given the question: [\texttt{What is the relationship between the white truck and this neighborhood?}]
what are the possible targets of the [\texttt{Video}] mainly mentioned or involved?\\
$<$/Question$>$\\

\vspace{2mm}
$\blacktriangleright$ \textbf{Output:}\\
The involved targets are [\texttt{the white truck}], [\texttt{the neighborhood}].
\end{tcolorbox}

\paragraph{$\blacktriangleright$ Step-2: }

The detailed prompt for Step-2 is shown as follows:
\begin{tcolorbox}[title=Step-2: Object Tracking,colback=gray!5,colframe=nmgray!75!black]
$\blacktriangleright$ \textbf{Input:}\\

$<$Question$>$\\ 
Provide the tracklet of involved [\texttt{the neighborhood}] and [\texttt{the white truck}] by outputting the corresponding partial            
expression in the [\texttt{STSG}].\\
$<$/Question$>$\\

$\blacktriangleright$ \textbf{Output:}\\
The partial STSG in tracking [\texttt{the neighborhood}] and [\texttt{the white truck}] is [Frame 1: {Objects: [{"car-1": [0.0,13.4,7.0,8.1]}, ...], Triplets: [("car-1","on the left", "street"), ...]}...].
\end{tcolorbox}

\paragraph{$\blacktriangleright$ Step-3:}

The detailed prompt for Step-3 is shown as follows:
\begin{tcolorbox}[title=Step-3: Action Analyzing,colback=gray!5,colframe=nmgray!75!black]

$\blacktriangleright$ \textbf{Input:}\\

$<$Question$>$\\
Combining all possible related commonsense, analyze the motion behavior based on the [\texttt{the white truck}] and [\texttt{the neighborhood}] and the neighbor scenes within [\texttt{STSG}].
Describing the action observations and implications.\\
$<$/Question$>$\\

$<$STSG$>$\\
Frame 1: \{\\
\hspace*{10pt}  Objects: [\{"car-1": [0.0,13.4,7.0,8.1]\}, ...], \\
\hspace*{10pt} Triplets: [("car-1", "on the left", "street"), ...]\\
\}\\
...\\
$<$/STSG$>$\\

$\blacktriangleright$ \textbf{Output:}\\
The two men are driving the white truck into a neighborhood, and pour the garbage from the roadside trash cans into the white truck. According to commonsense, the white car is used for collecting rubbish...

\end{tcolorbox}

\paragraph{$\blacktriangleright$ Step-4:}

When the raw question is an open-ended QA, we consider prompting the model to output multiple distinct optional answers, such that we unify all QA problems into the Multi-choice type:
\begin{tcolorbox}[title=Step-4-Pre: Transforming Open-ended Question Answering into Multi-choice one,colback=gray!5,colframe=nmgray!75!black]

$\blacktriangleright$ \textbf{Input:}\\
$<$Question$>$\\
For the question [\texttt{What is the relationship between the white truck and this neighborhood?}], 
please based on the action's [\texttt{The two men are driving the white truck into a neighborhood...}] combined with commonsense, output 4 distinct optional answers with the rationality score of this answer with a 1-10 scale.
$<$/Question$>$\\

$\blacktriangleright$ \textbf{Output:}\\
Answer A: While the white truck is indeed moving through ... but rather the collection of garbage ... \\
Answer B: ...\\
\end{tcolorbox}

Given the multiple-choice question, we first prompt the model to score its likelihood (from 1 to 10) in conjunction with commonsense knowledge, and provide a corresponding rationale for each candidate answer.
Then, we consider a ranking mechanism to determine the final answer.
\begin{tcolorbox}[title=Step-4-A: Multi-choice Question Answering via Ranking,colback=gray!5,colframe=nmgray!75!black]

$\blacktriangleright$ \textbf{Input:}\\
$<$Question for Answer A$>$\\
For the question [\texttt{What is the relationship between the white truck and this neighborhood? A. Transportation  B. Buildings  C. Clean Services  D. Entertainment Facilities}], given a candidate answer [\texttt{A. Transportation}], please based on the action's [\texttt{The two men are driving the white truck into a neighborhood...}] combined with commonsense, score the rationality of this answer with a 1-10 scale, and also output the rationale.\\
$<$/Question for Answer A$>$\\

\vspace{2mm}
$<$Question for Answer B$>$\\
For the question [\texttt{What is the relationship between the white truck and this neighborhood? A. Transportation  B. Buildings  C. Clean Services  D. Entertainment Facilities}], given a candidate answer [\texttt{B. Buildings}], please based on the action's [\texttt{The two men are driving the white truck into a neighborhood...}] combined with commonsense, score the rationality of this answer with a 1-10 scale, and also output the rationale.\\
$<$/Question for Answer B$>$\\

\vspace{2mm}
$<$Question for Answer C$>$\\
For the question [\texttt{What is the relationship between the white truck and this neighborhood? A. Transportation  B. Buildings  C. Clean Services  D. Entertainment Facilities}], given a candidate answer [\texttt{C. Clean Services}], please based on the action's [\texttt{The two men are driving the white truck into a neighborhood...}] combined with commonsense, score the rationality of this answer with a 1-10 scale, and also output the rationale.\\
$<$/Question for Answer C$>$\\

\vspace{2mm}
$<$Question for Answer D$>$\\
For the question [\texttt{What is the relationship between the white truck and this neighborhood? A. Transportation  B. Buildings  C. Clean Services  D. Entertainment Facilities}], given a candidate answer [\texttt{D. Entertainment Facilities}], please based on the action's [\texttt{The two men are driving the white truck into a neighborhood...}] combined with commonsense, score the rationality of this answer with a 1-10 scale, and also output the rationale.\\
$<$/Question for Answer D$>$\\

$\blacktriangleright$ \textbf{Output:}\\
Based on the observations and commonsense, the rationality of the answer [\texttt{A. Transportation}] is 5. While the white truck is indeed moving through ... but rather the collection of garbage ... \\

\end{tcolorbox}

\begin{tcolorbox}[title=Step-4-B: Multi-choice Question Answering via Ranking,colback=gray!5,colframe=nmgray!75!black]

$\blacktriangleright$ \textbf{Input:}\\
$<$Question for Ranking$>$\\
Now, we know the rationale score of the answer [\texttt{A. Transportation}] is 5, ... Please rank the candidate the answer based on the rationale score of each candidate's answer.\\
$<$/Question for Ranking$>$\\

$\blacktriangleright$ \textbf{Output:}\\
 Based on the rationality, the ranking results in descending order is [\texttt{C. Clean Service}] [\texttt{A. Transportation}] [\texttt{D. Entertainment Facilities}]  [\texttt{B. Buildings}]. So, the final answer is [\texttt{C. Clean Service}].
\vspace{2mm}
\end{tcolorbox}

\paragraph{ $\blacktriangleright$ Step-5:}

The detailed prompt for Step-5 is shown as follows:
\begin{tcolorbox}[title=Step-5: Answer Verification,colback=gray!5,colframe=nmgray!75!black]
$\blacktriangleright$ \textbf{Input:}\\
$<$Video$>$ \videologo $<$/Video$>$\\

$<$Question$>$\\
Given the [\texttt{Video}], and the raw question [\texttt{What is the relationship between the white truck and this neighborhood? A. Transportation  B. Buildings  C. Clean Services  D. Entertainment Facilities}], now you need to verify the previous answer by\\
\hspace*{10pt}  1) checking the pixel grounding information if the answer [\texttt{C. Clean Service}] aligns with the facts presented in the video from a perception standpoint; \\
\hspace*{10pt}  2) determining from a cognition perspective if the commonsense implications inherent in the answer [\texttt{C. Clean Service}] contradict any of the main [\texttt{Observations}] inferred in the 3-$rd$ reasoning step.\\
Output the verification result with rationale.\\
$<$/Question$>$\\

$\blacktriangleright$ \textbf{Output:}\\
Certainly. Let's verify the answer [\texttt{C. Clean Service}]:\\
\hspace*{10pt}1. Grounding information: the collection action of garbage, which aligns ... \\
\hspace*{10pt}2. Commonsense verification: driving a white truck into a neighborhood is associated with waste management ...  \\
Overall, the answer [\texttt{C. Clean Service}] is correct.
\end{tcolorbox}

Note that the above demonstration of the entire prompting process is just one example. 
In practice, we simply need to replace the example instance with any other QA instance. 
It is also worth noting that the VoT system is not limited to video QA; any other video tasks and scenarios requiring reasoning can be adapted to this framework by transforming the original task into a QA format.

\subsection{Implementations}

MotionEpic uses the Vicuna-7B (v1.5)\footnote{\url{https://huggingface.co/lmsys/vicuna-7b-v1.5}} as the backbone LLM.
We adopt the ViT-L/14\footnote{\url{https://huggingface.co/sentence-transformers/clip-ViT-L-14}} as the video encoder, and use the Q-Former\footnote{\url{https://huggingface.co/spaces/Salesforce/BLIP2}} as the projector.
All the modules take the default configurations without much modification.
For our Recurrent Graph Transformer, we take a 6-layer architecture with 768-d hidden sizes.
The text tokenizer is sourced from LLaMA, with approximately 32,000 classes.
For each video, we uniformly sample certain frames with a sampling rate of 8 fps for fine-grained reasoning. 
We note that too large sampling rate introduces noises (i.e., redundant frames) and huge computation cost, while too small one will cause important information loss.
Here we use the 8 fps, as in our preliminary study we verified that it helps achieve the best trade-off.
For the fine-tuning setting of end tasks, we will tune the MotionEpic based on the training set using the setting as prior baselines, i.e., data split and evaluation methods.
For the zero-shot setting, we will directly perform video QA without using the in-domain training set.

\section{More Qualitative Visualizations}

Finally, we provide two sets of cases for qualitative analyses. 
We observe that different Video QA datasets exhibit varying biases. 
Some datasets lean more towards content recognition, relying heavily on perceptual abilities without necessitating much cognitive understanding; others are more inclined towards cognitive-level comprehension, such as physical, cultural or humanities knowledge, where the video content itself is relatively straightforward. 
We consider cases from both these perspectives.

\begin{figure*}[h]
\centering
\includegraphics[width=1\textwidth]{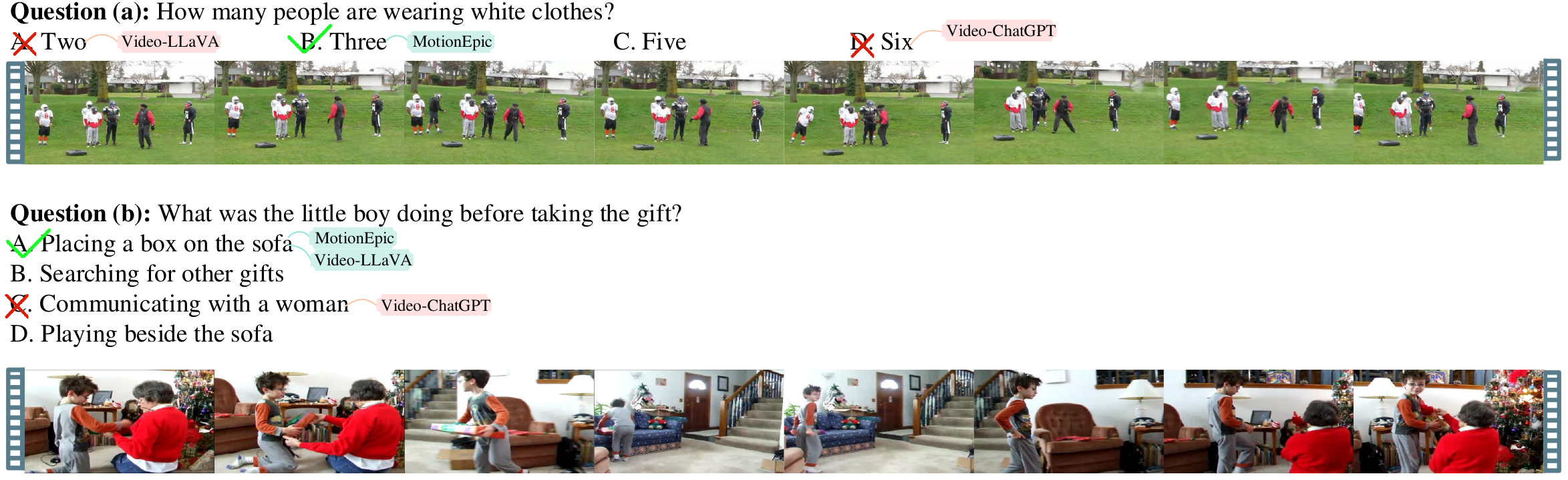}
\caption{
Qualitative examples of perception-level reasoning. The correct answer is marked with a green checkmark, and the wrong answer is marked with a red cross.
}
\label{fig:app-case-perp}
\end{figure*}

Fig. \ref{fig:app-case-perp} presents two sets of QA cases at the video perception level. 
For the first question, which requires counting the number of people in the video, it is observed that both baselines provided incorrect answers. However, thanks to our MotionEpic's utilization of the STSG structured representation, it can accurately ground the number of objects, thereby providing the correct result. 
In the second case, a straightforward understanding of the temporal information in the video suffices to answer the question. 
It is shown that both MotionEpic and Video-LLaVA answered correctly.

\begin{figure*}[h]
\centering
\includegraphics[width=1\textwidth]{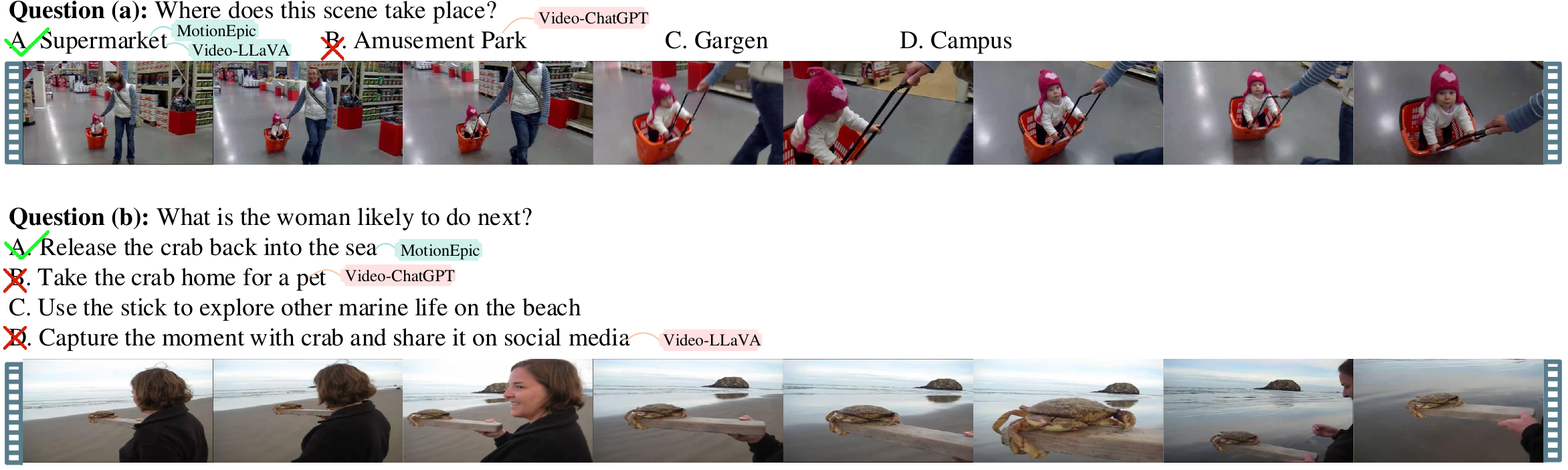}
\caption{
Qualitative examples of cognitive-level reasoning.
}
\label{fig:app-case-cog}
\end{figure*}

Fig. \ref{fig:app-case-cog} showcases two cases at the cognitive level. 
For the first case, the question ``Where does this scene take place?'' can be answered by understanding the scene's content and combining it with common sense to conclude: a supermarket. 
For the second case, merely observing the video content ``a woman holds a crab with a stick'' makes it challenging to grasp the implicit intention. 
However, integrating some cultural commonsense, it can be understood that the girl is releasing the crab back into the sea.

\end{document}